\documentclass[twoside,leqno,twocolumn]{article}

\usepackage[letterpaper]{geometry}

\usepackage{ltexpprt}
\usepackage{hyperref}
\usepackage{graphicx}
\usepackage{cite}
\usepackage{algorithm}
\usepackage[noend]{algorithmic}
\usepackage{amsfonts,amssymb}
\usepackage{tikz}
\usepackage{multirow}

\usepackage{marvosym}
\begin{document}

\newcommand\relatedversion{}

\title{\Large Prompt Based Tri-Channel Graph Convolution Neural Network for Aspect Sentiment Triplet Extraction\relatedversion}

\author{Kun Peng\thanks{Institute of Information Engineering, Chinese Academy of Sciences, Beijing, China. \{pengkun, jianglei\}@iie.ac.cn} \thanks{School of Cyberspace Security, University of Chinese Academy of Sciences, Beijing, China. }
\and  Lei Jiang$^*$\textsuperscript{\Letter}
\and Hao Peng\thanks{School of Cyber Science and Technology, Beihang University, Beijing, China. penghao@buaa.edu.cn} \textsuperscript{\Letter}
\and Rui Liu$^{*\dagger}$
\and Zhengtao Yu\thanks{Faculty of Information Engineering and Automation, Kunming University of Science and Technology, Kunming, China.}
\and Jiaqian Ren$^{*\dagger}$
\and Zhifeng Hao\thanks{College of Science, University of Shantou, Shantou, China.}
\and Philip S.Yu\thanks{University of Illinois at Chicago, Department of Computer Science, Chicago, IL, USA.} 
}

\date{}

\maketitle


\fancyfoot[R]{\scriptsize{Copyright \textcopyright\ 2024 by SIAM\\
Unauthorized reproduction of this article is prohibited}}





\begin{abstract} \small\baselineskip=9pt 
Aspect Sentiment Triplet Extraction (ASTE) is an emerging task to extract a given sentence's triplets, which consist of aspects, opinions, and sentiments.
Recent studies tend to address this task with a table-filling paradigm, wherein word relations are encoded in a two-dimensional table, and the process involves clarifying all the individual cells to extract triples.
However, these studies ignore the deep interaction between neighbor cells, which we find quite helpful for accurate extraction. 
To this end, we propose a novel model for the ASTE task, called \textbf{P}rompt-based \textbf{T}ri-Channel \textbf{G}raph \textbf{C}onvolution Neural \textbf{N}etwork (PT-GCN), which converts the relation table into a graph to explore more comprehensive relational information. 
Specifically, we treat the original table cells as nodes and utilize a prompt attention score computation module to determine the edges' weights. This enables us to construct a target-aware grid-like graph to enhance the overall extraction process.
After that, a triple-channel convolution module is conducted to extract precise sentiment knowledge.
Extensive experiments on the benchmark datasets show that our model achieves state-of-the-art performance.
The code is available at https://github.com/KunPunCN/PT-GCN.

\textbf{Keywords}: Text Mining, Aspect Sentiment Triplet Extraction, Prompt Learning

\end{abstract}

\section{Introduction}
\label{sec:intro}
\begin{figure}[t]
    \centering
    \includegraphics[width=0.395\textwidth]{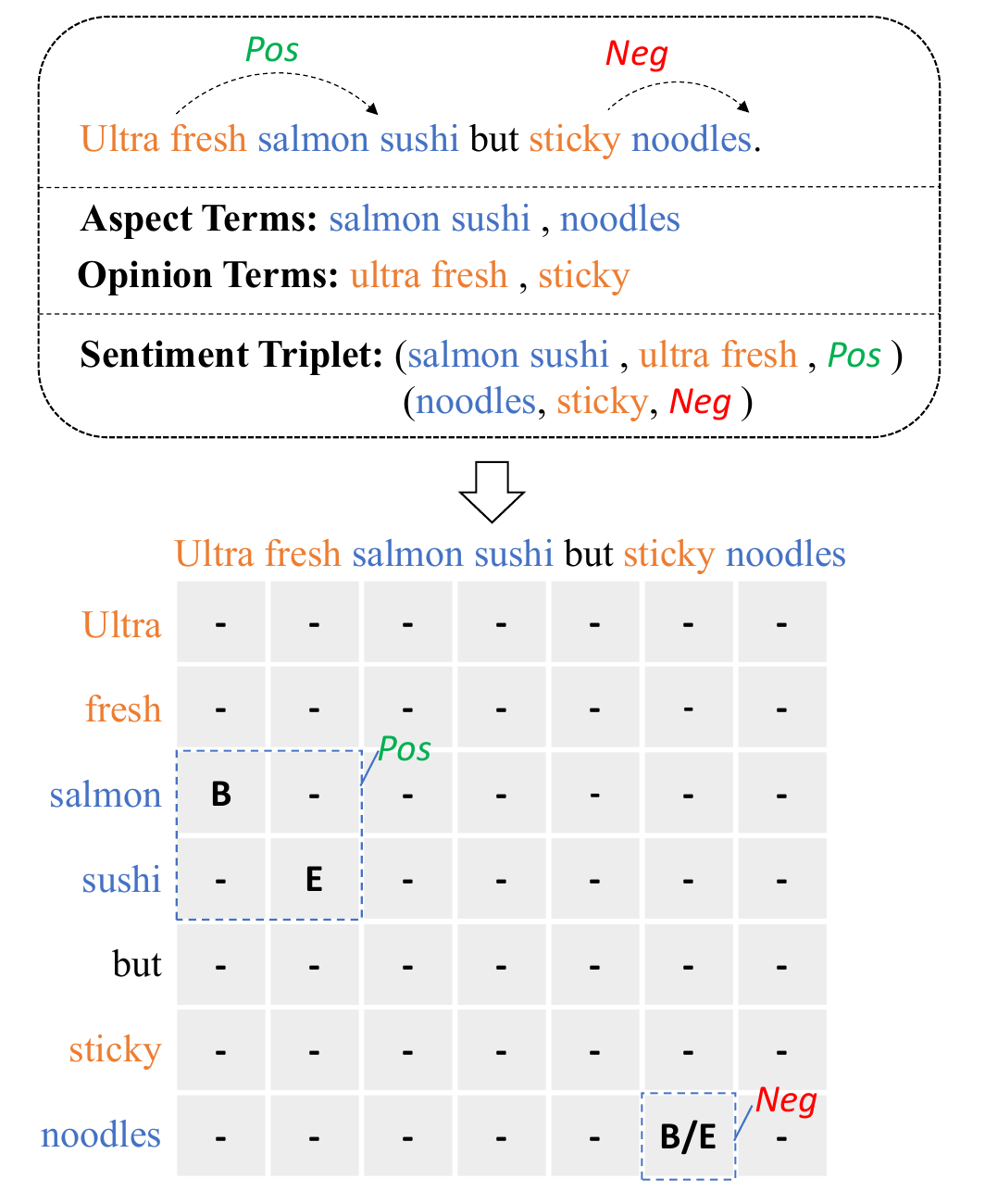}
    \vspace{-0.5cm}
    \caption{An example of the ASTE task and table filling scheme. In the relation table, ``B'' and ``E'' label the vertex of the regions borderline, and then the restricted regions are classified into sentiment categories.}
    \label{fig:1}
\end{figure}

Sentiment analysis is an important research direction in natural language processing. 
It is used for opinion mining of commentary texts such as social news and user comments \cite{2014Sentiment}.
Recent studies have continuously proposed tasks such as Aspect-based Sentiment Analysis \cite{pontiki-etal-2014-semeval,zhao2023rdgcn} and Aspect-category Sentiment Detection \cite{2020Target}, aiming to extract fine-grained sentiment elements.
To further explore fine-grained extraction, the Aspect Sentiment Triplet Extraction (ASTE) task recently attracted more attention from researchers. 
As shown in the upper part of Figure \ref{fig:1}, the ASTE task aims to extract the triplet: (aspect term, opinion term, sentiment polarity). 
In this example, the aspect and opinion terms are highlighted in blue and orange, respectively.
The phrase ``\textcolor{orange}{ultra fresh}'' serves as an opinion term for ``\textcolor{blue}{salmon sushi}'', expressing a positive sentiment. 
``\textcolor{orange}{sticky}'' is an opinion term for ``\textcolor{blue}{noodles}'', indicating a negative sentiment.
This compound task requires not only extracting the correct terms but also identifying the sentiment relation between them, making it more challenging.

There are two paradigms in existing research called pipeline and joint extraction.
The pipeline method \cite{DBLP:conf/aaai/PengXBHLS20,xu-etal-2021-learning,DBLP:conf/aaai/MaoSYC21} decomposes the task into multiple subtasks and devises different modules to address them step by step, inevitably introducing cascading errors between modules.
The joint extraction method exploits the association between subtasks, including Sequence Tagging \cite{xu-etal-2020-position}, Generative \cite{yan-etal-2021-unified,zhang-etal-2021-towards-generative}, Table Filling \cite{2016Table,wu-etal-2020-grid,Chen2022EnhancedMG,zhang-etal-2022-boundary}, and other methods.
Table Filling performs significantly well among the joint extraction methods due to its well-designed framework. 
As shown in the lower part of Figure \ref{fig:1}, it first encodes word relations on a two-dimensional table, one for aspect and another for opinion, and then classifies all the table cells to extract triplets. 
Recent table-filling methods focus on learning better word-level representations for cells.
GTS \cite{wu-etal-2020-grid} encodes word representations with typical neural networks, i.e., BERT \cite{DBLP:conf/nips/VaswaniSPUJGKP17}, and simply concatenates them together to form a table. 
EM-GCN \cite{Chen2022EnhancedMG} enhances word representations through syntactic dependency parsing. 
Dual-Encoder \cite{jing-etal-2021-seeking} uses a sequence encoder with a pair encoder to fully learn word representation.

Although these efforts are valuable, they simply exploit independent word-level information in each cell and ignore interactions between neighboring cells.
For example, the cell$_{ij}$ represents the relations between the $i$-th and $j$-th words. 
Without a vision of its surrounding cells' semantics, cell$_{ij}$ lacks span-level and boundary information, which leads to errors in the table decoding process.
To this end, we propose an effective end-to-end model called \textbf{P}rompt-based \textbf{T}riple-Channel \textbf{G}raph \textbf{C}onvolution Neural \textbf{N}etwork (\textbf{PT-GCN}). 
Instead of encoding word relations in a two-dimensional table, we further transform the table into a target-aware relation graph to contain more contextual neighbor information.
Specifically, first, we view the table as a tensor map, treating each cell as a node, its vector as the initial node representation. 
To obtain the weight of each edge in the graph, we propose a prompt attention score computation module that calculates the attention score for each word and automatically learns edge weights.
This table-to-graph transformation introduces more contextual information.
Then, to learn precise relation information from different sentiment perspectives, we propose a triple-channel parallel convolution module based on multiplex sentiment prompts.
Finally, The learned representations are integrated for table decoding.

Our contributions are summarized as follows: 

1) We propose an effective PT-GCN model for the ASTE task. 
It can fully learn the relations of both word-level and span-level.

2) We propose a prompt-based table-to-graph transformation method.
With the edge weights learned from a prompt attention module, the model will better perceive node relations.
The relation graph is ingeniously constructed through prompt learning without relying on external knowledge.

3) We propose a novel parallel convolution module, which fully learns word relations from different sentiment perspectives.

4) Extensive experiments conducted on benchmark datasets have demonstrated that our model achieves the best publicly available results on the ASTE task.

\section{Related Work}
\textbf{ASTE.} \quad The ASTE task has gained traction in recent years. ASTE has two research lines, including pipeline and joint extraction.
As an emerging compound task, it can be decomposed into a series of subtasks, such as Aspect Term Extraction (ATE), Opinion Term Extraction (OTE) and Aspect Sentiment Classification (ASC) \cite{chen-qian-2019-transfer}.
Most recently, Peng \cite{DBLP:conf/aaai/PengXBHLS20} first proposed the ASTE task and addressed it with a two-stage pipeline method. 
They solved these three subtasks separately in the first stage and paired the triplets using a classifier in the second stage.
Both Chen \cite{DBLP:conf/aaai/ChenWLW21} and Mao \cite{DBLP:conf/aaai/MaoSYC21} constructed the ASTE task as a multi-step machine reading comprehension (MRC) problem.
However, these pipeline methods are limited by error propagation and its cumbersome process.

Another research line is joint extraction.
Xu \cite{xu-etal-2020-position} designed a position-aware sequence tagging approach to capture the rich relations among the sentiment elements. 
Yan \cite{yan-etal-2021-unified} used the generative model BART and a pointer network to indicate the target's location. 
Zhang \cite{zhang-etal-2021-towards-generative} designed two types of paradigms, named annotation-style and extraction-style modeling, tackling several sentiment analysis tasks in a unified generative framework.
Xu \cite{xu-etal-2021-learning} proposed a span-level approach that explicitly considers the interaction between the whole spans of aspects and opinions. 
Among these joint extraction methods, Table Filling is an attractive approach.
Wu \cite{wu-etal-2020-grid} first transformed ASTE into a unified table-filling task and proposed the Grid Tagging Scheme (GTS) to address it.
To further explore this method, Jing \cite{jing-etal-2021-seeking} introduced a dual-encoder to learn subtasks' differences.
Chen \cite{Chen2022EnhancedMG} proposed a multi-channel GCN model, which fully utilizes word relations based on semantic dependency trees and the multi-channel graph.
Zhang \cite{zhang-etal-2022-boundary} proposed a boundary detection model, which can fully exploit both word-to-word and relation-to-relation interactions.

\begin{figure*}[t]
    \centering
    \includegraphics[width=0.95\textwidth]{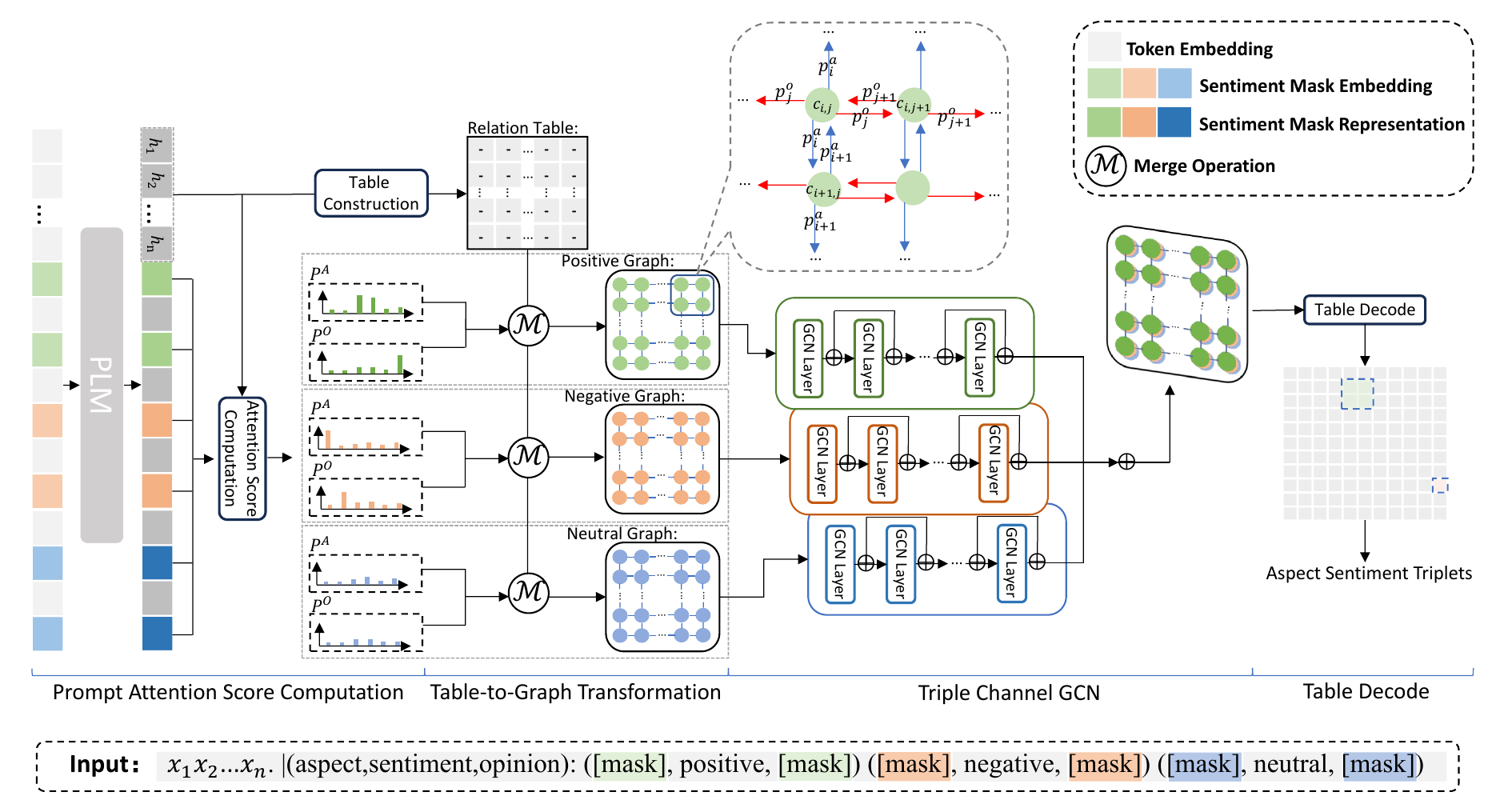}
    \caption{The overview of PT-GCN. Merge Operation means merging the relation table and prompt scores to construct the sentiment graph.}
    \label{fig_main}
\end{figure*}

\noindent \textbf{Prompt-tuning.} \quad Prompt-tuning is a fine-tuning paradigm proposed in GPT-3 \cite{Brown2020LanguageMA} and it gets amazing results in few-shot or zero-shot scenarios.
It designs a template indicating task information for input, which makes the downstream tasks more suitable for the model training process.
Prompt-tuning helps pre-trained language models better mine knowledge \cite{petroni-etal-2019-language}, and make progress on many NLP tasks, such as Text Classification \cite{hu-etal-2022-knowledgeable}, Natural Language Inference \cite{qi-etal-2022-enhancing} and Sentiment Analysis \cite{wu-shi-2022-adversarial}.
Unlike these tasks, the ASTE task is composite and more complex. 
Applying prompt learning on ASTE is worth exploring.
Li \cite{li2021sentiprompt} explored using prompt learning in the ASTE task with a generative architecture.

\section{Approach}
\label{sec:approach}
The model's architecture is shown in Figure \ref{fig_main}.
It consists of four parts: Prompt Attention Score Computation, Table-to-Graph Transformation, Triple Channel GCN, and Table Decoding.
We will discuss each part according to the model process in the rest of this section.

\subsection{Task Formulation.}
Given a sentence of \emph{n} words \(X=\left \{ x_{1},x_{2}...x_{n} \right \}\), the ASTE task aims at extracting a set of triplets \(T=\{ t_{i} \}_{i=1}^I\) from \(X\), where the \(i\)-th triplet \(t_{i}=(a_{i},o_{i},l_{i})\). \(a_{i}\) and \(o_{i}\) denote the aspect term and opinion term in \(X\), respectively. \(l_{i}\) is the sentiment label in \( \{ \emph{Pos}, \emph{Neg}, \emph{Neu} \}\). 
As shown in the lower part of Figure \ref{fig:1}, we encode the word-pair relation representations in a two-dimensional table and extract the triplets in a region detection paradigm.
Two entity labels \( \{ \emph{B}, \emph{E} \}\) on the non-diagonal are used for tagging the beginning and end positions of the aspect-opinion pair, respectively.
After that, we extract all detected regions that are restricted by the entity labels, and
use three sentiment labels \( \{ \emph{Pos}, \emph{Neg}, \emph{Neu} \}\) to classify the sentiment polarities of the candidate triplets.

\subsection{Prompt Attention Score Computation.}
\begin{figure}[t]
    \centering
    \includegraphics[width=0.48\textwidth]{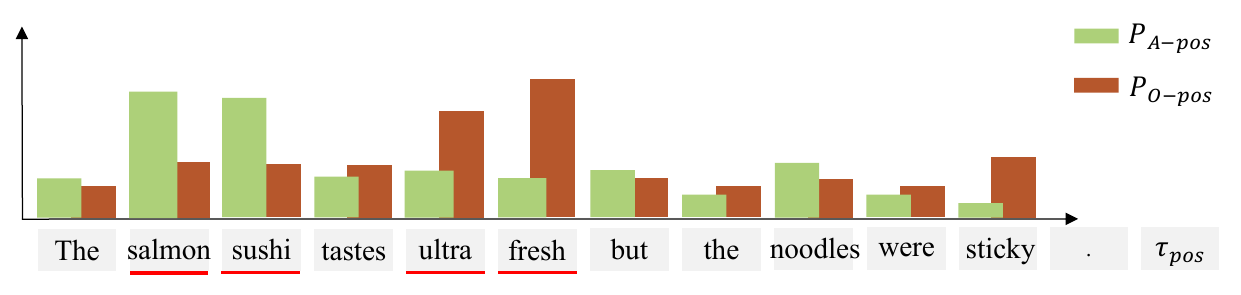}
    \vspace{-0.3cm}
    \caption{An example of attention score for positive sentiment.}
    \label{fig:4}
\end{figure}

As shown in the bottom of Figure \ref{fig_main}, given a sentence \(X=\left \{ x_{1},x_{2}, ..., x_{n} \right \}\), we concatenate it with a manually designed template $\mathcal{T}$. For each sentiment polarity, there are two mask slots for aspect and opinion. Then we choose BERT \cite{DBLP:conf/nips/VaswaniSPUJGKP17} as the language encoder to obtain the representations of the input sentence. The encoding process is formulated as,
\begin{equation}
[H|H_\mathcal{T}] = BERT([X|\mathcal{T}]), \nonumber
\end{equation}
where $[H|H_\mathcal{T}]\in\mathbb{R}^{n\times d}$ is the last hidden layer of the encoder output. $d$ is the vector dimension. 
$H_\mathcal{T}$ contains six representations corresponding to the ``\emph{[MASK]}'' slots: $[\tau_{pos}^a, \tau_{pos}^o, \tau_{neg}^a, \tau_{neg}^o, \tau_{neu}^a, \tau_{neu}^o]$,
where $\tau \in\mathbb{R}^{6\times d}$ contains information about aspects and opinions that correspond to different sentiments: positive (pos), negative (neg), and neutral (neu).
To indicate the positions of aspects and opinions in the sentence, we calculate the attention score $P \in \mathbb{R}^{6\times n}$ using $\tau$ and $H$:
\begin{equation}
P = softmax(\tau W H^T),\label{3.2}
\end{equation}
where $W \in\mathbb{R}^{d\times d}$ is a learnable parameter. 
As an example shown in Figure \ref{fig:4}, when considering positive sentiment, $P^{A}_{pos} $ has the highest probability at \emph{salmon sushi}, and $P^{O}_{pos} $ has the highest probability at \emph{ultra fresh}.

\subsection{Table-to-Graph Transformation.}
\label{tabletograph}
%

To capture the word-level relations, we constructed a relation table $C$ using $H$. 
For a cell $c_{ij} \in C$, it represents the relation between the $i$-th and $j$-th words of the given sentence.
Inspired by BDTF \cite{zhang-etal-2022-boundary}, we use a maxpooling operation on the sequence between $h_i \in \mathbb{R}^{d}$ and $h_j \in \mathbb{R}^{d}$, which helps to capture more contextual information.
In addition, a tensor-based operation \cite{socher-etal-2013-recursive} is used to capture the word interactions, which has proven effective.
By concatenating various forms of computation, $c_{ij}$ can fully learn relations information.
The calculation process is as follows:
\begin{equation}
\hat{h}_{ij} = h_i \oplus h_j \oplus pooling(h_{i:j}) \oplus (h_i)^TW_1h_j,\label{eq2.6}
\end{equation}
\begin{equation}
c_{ij} = \sigma (W_2\hat{h}_{ij}),\label{eq3}
\end{equation}
where $W_1\in \mathbb{R}^{d\times t\times d}$ and $W_2\in \mathbb{R}^{d\times (3d+t)}$ are trainable parameters. $\sigma$ is an activation function (e.g., ReLU). $\oplus$ denotes the concatenation operation. 
$h_{i:j} (i \leq j)$ is a sequence of $[h_i, h_{i+1}, ..., h_j]$. 
$(h_i)^TW_1h_j$ is the tensor-based operation.

As we can observe, the relation table $C$ lacks sufficient sentiment relation information between adjacent nodes.
For example, in Figure \ref{fig:1}, the relation between nodes ``noodles-were'' and ``noodles-sticky'', has a huge difference with the relation between nodes ``were-were'' and ``were-sticky''.
Therefore, we wish to add more sentiment information between nodes in the relation table.
To transform the relation table into a graph, we treat each cell $c_{ij}$ in the table as a node in the graph, and define ${r}^{ij}_{i(j+1)}$ as the relation arc connecting node $c_{ij}$ to node $c_{i(j+1)}$.
%
As depicted in the upper right corner of Figure \ref{fig_main}, each node in the table (excluding edge nodes) has four outgoing arcs and four incoming arcs.
The higher the probability of a node being a term word-pair, the stronger its impact on surrounding nodes.
For example, in the positive relation graph, we can define the weights of the outgoing arcs as follows:
\begin{equation}
r^{ij}_{i(j+1)} = r^{ij}_{i(j-1)} = p^{pos_a}_j,
\end{equation}
\begin{equation}
r^{ij}_{(i+1)j} = r^{ij}_{(i-1)j} = p^{pos_o}_i,
\end{equation}
where $p^{pos_a}_j, p^{pos_o}_i \in P$ are prompt scores for the positive aspect slot and positive opinion slot.
With table nodes $C$ and arcs $R_{pos} = \{r_{11},r_{12}...,r_{nn}\}$, we define a directed positive graph $G_{pos}=\{C, R_{pos}\}$. In the same way, we can define the negative graph $G_{neg}$ and the neutral graph $G_{neg}$.

\subsection{Triple Channel GCN.}
\label{triple-gcn}
To model varied relations between word-pair nodes, we apply a Graph Convolutional Neural Network (GCN) \cite{kipf2017semisupervised} to aggregate relevant information from adjacent nodes.
Because the grid-like graph has local structural consistency, we can simplify the convolution process and accelerate calculations.
Specifically, given the $(l-1)$-th layer input $G^{(l-1)}$, the $l$-th layer GCN produces $G^{l}\in \mathbb{R}^{n\times n\times d}$ by:
\begin{equation}
g^l_{ij} = \sigma ( W_g^{l} ( g^{l-1}_{ij} + \sum_{k \in \{(i\pm1)j, i(j\pm1)\} } r^k_{ij} g^{l-1}_k ) ),
\end{equation}
where $W_g^{l}\in \mathbb{R}^{d\times d}$ is a trainable parameter. 
The first layer $g^0_{ij}$ is initialized to $c_{ij}$.
After the multi-layer GCN, we input the last hidden layer representation into a linear layer to obtain the final node representation $\tilde{C}$.

There exist three sentiment graphs with the same nodes but have significant differences in edge weights, each capturing different sentiment relations.
We conduct parallel graph convolution operations on the three sentiment graphs, each channel focuses on learning single precise sentiment relation.
%
%
The final table $C'$ is produced by:
\begin{equation}
C' = \tilde{C}_{pos} \oplus \tilde{C}_{neg} \oplus \tilde{C}_{neu},\label{eq3.13}
\end{equation}
where $\tilde{C}_{pos}$, $\tilde{C}_{neg}$, $\tilde{C}_{neu}$ are representations corresponding to graphs $G_{pos}$, $G_{neg}$ and $G_{neu}$, respectively.

\subsection{Table Decoding.}
\label{tabledecoding}
Similar to BDTF \cite{zhang-etal-2022-boundary}, our decoding process contains two steps: detection and classification.

\textbf{Detection:}  To detect the target regions, we input $C'$ into two parallel linear layers, followed by a sigmoid function, to obtain two entity score tables $S^S \in \mathbb{R}^{n\times n}$ and $S^E \in \mathbb{R}^{n\times n}$. 
Each $s_{ij}^S$ in $S^S$ represents the score of the upper left corner of the target region. $s_{ij}^E$ in $S^E$ represents the score of the lower right corner.
After that, we use a top-k strategy to get the candidate entities.
Given a score table $S$ and a span pruning threshold $k$, we first sort the scores and then take the top $k\%$ scores' position (a tuple like $(i, j)$) as candidates. 
Corresponding to $S^S$ and $S^E$, the two candidate sets are $S$ and $E$. 

\textbf{Classification: }  Firstly, we traverse the two candidate sets $S$ and $E$ to form the candidate regions.
For two position tuples $(a, b)$ in $S$ and $(c, d)$ in $E$, the candidate region is the restricted rectangle $C_{a:c, b:d} \in C' (a \leq c, b \leq d)$.
The representation of the candidate region is produced by:
\begin{equation}
    R = C_{a,b} \oplus C_{c,d} \oplus maxpooling(C_{a:c,b:d}). \label{eq3.16}
\end{equation}
It is concatenated by the two vertices and a maxpooling of the region $C_{a:c, b:d}$.
Finally, we input $R$ into a linear layer, followed by a softmax function, to obtain the region's sentiment classification result $s_{abcd}^*$.
The value of $s_{abcd}^* \in \{0, 1, 2, 3\}$ corresponds to label in \{\emph{Padding}, \emph{Pos}, \emph{Neg}, \emph{Neu}\}.
%
We drop the \emph{Padding} region and finally obtain a sentiment triplet:
$t = ((a, c), (b, d), \mathcal{F}(s_{abcd}^*))$,
where $\mathcal{F}$ is a label mapping function.

\subsection{Training Loss.}
Given the ground truth entity label $y_{ij}^S, y_{ij}^E \in \{0, 1\}$, the entity loss is formulated as a sum of two cross-entropy losses: $\mathcal{L}_{1}=\mathcal{L}_s+\mathcal{L}_e$, where
\begin{equation}
    \mathcal{L}_s=-\sum_{i=1}^n\sum_{j=1}^n \sum_{k\in [0,1]} \mathbb I(y_{ij}^S=k)log(s_{ij}^S),
\end{equation}
\begin{equation}
    \mathcal{L}_e=-\sum_{i=1}^n\sum_{j=1}^n \sum_{k\in [0,1]} \mathbb I(y_{ij}^E=k)log(s_{ij}^E).
\end{equation}
$\mathbb I(\cdot)$ is the indicator function. Given the sentiment label $y^* \in \{0,1,2,3\}$, the sentiment loss is calculated by:
\begin{equation}
    \mathcal{L}_{2}=-\sum_{i=1}^m\sum_{k\in [0,1,2]} \mathbb I(y_i^*=k)log(s_{abcd}^*),
\end{equation}
where $m$ is the number of candidate regions. 
Our final training goal is to minimize $\mathcal{L}$:
\begin{equation}
    \mathcal{L}=\alpha\mathcal{L}_{1} + (1-\alpha)\mathcal{L}_{2},
\end{equation}
where $\alpha>0$ is a hyperparameter.

\section{Experiments} 
\subsection{Datasets.}
Following previous studies \cite{Chen2022EnhancedMG}, we evaluate our model on two public datasets\footnote{https://github.com/xuuuluuu/SemEval-Triplet-data} that are derived from the SemEval ABSA Challenges \cite{pontiki-etal-2014-semeval}.
The first dataset $\mathcal{D}_1$ is released by \cite{DBLP:conf/aaai/PengXBHLS20}. The second dataset $\mathcal{D}_2$ is a revised version of $\mathcal{D}_1$, which comes from \cite{xu-etal-2020-position}. 
These two datasets both include three restaurant datasets and one laptop dataset.
The detailed statistics are shown in Table \ref{tab:statistics}, where \#S denotes the number of sentences, and \#T denotes the number of triplets.

\begin{table}[h]\small
\caption{Statistics of datasets.}
\label{tab:statistics}
\centering
\tabcolsep=3pt

\scalebox{1}{
\begin{tabular}{lllllllllll}
\hline
\multicolumn{2}{c}{\multirow{2}{*}{\textbf{Dataset}}} 
    & \multicolumn{2}{c}{14res} & \multicolumn{2}{c}{14lap} & \multicolumn{2}{c}{15res} & \multicolumn{2}{c}{16res} \\ \cline{3-10} 
    & & \#S        & \#T          & \#S         & \#T         & \#S        & \#T          & \#S        & \#T          \\ \hline
\multirow{3}{*}{$\mathcal{D}_1$}              
    & train           & 1,259       & 2,356             & 899        & 1,452     & 603        & 1,038        & 863        & 1,421        \\
    & dev             & 315         & 580               & 225        & 383       & 151        & 239          & 216        & 348          \\
    & test        & 493         & 1,008                & 332        & 547        & 325        & 493          & 328        & 525          \\ \hline
\multirow{3}{*}{$\mathcal{D}_2$}              
    & train            & 1,266       & 2,338               & 219        & 346        & 605        & 1,013        & 857        & 1,394        \\
    & dev           & 310         & 577              & 219        & 346          & 148        & 249          & 210        & 339          \\
    & test            & 492         & 994              & 328        & 543        & 322        & 485          & 326        & 514          \\ \hline
\end{tabular}
}
\end{table}

\begin{table*}[h]
\centering
\caption{Results on $\mathcal{D}_1$. All baseline results are from the original papers.}
\tabcolsep=1.7pt
\scalebox{1}{
\begin{tabular}{lllllllllllll|c}
\hline
\multicolumn{1}{c}{\multirow{2}{*}{\textbf{Model}}}
    &  \multicolumn{3}{c}{14res}      & \multicolumn{3}{c}{14lap($\alpha=0.4$)}      & \multicolumn{3}{c}{15res}           & \multicolumn{3}{c|}{16res($\alpha=0.6$)} & \multirow{2}{*}{Avg-F1} \\  \cline{2-13}
    & P.    & R.    & F1             & P.    & R.    & F1             & P.    & R.    & F1             & P.    & R.    & F1 &{}      \\ \hline
\multicolumn{1}{l}{Peng+IOG}                           &58.89   &60.41   &59.64            &48.62   &45.52  &47.02           &51.70   &46.04  &48.71           &59.25   &58.09   &58.67 & 53.51 \\
\multicolumn{1}{l}{IMN+IOG}                        &59.57   &63.88   &61.65          &49.21  &46.23  &47.68          &55.24   &52.33   &53.75           &-  &-  &-  & -    \\
\multicolumn{1}{l}{BMRC}                        & - & - & 70.01          & - & - & 57.83          & - & - & 58.74          & - & - & 67.49 & 63.52 \\
\hline
\multicolumn{1}{l}{Dual-MRC}                   & 71.55 & 69.14 & 70.30          & 57.39 & 53.88 & 55.58          & 63.78 & 51.87 & 57.21          & 68.60 & 66.24 & 67.40   &62.62       \\
\multicolumn{1}{l}{BART-ABSA}                   & - & - & 72.46          & - & - & 57.59          & - & - & 60.11          & - & - & 69.98    &65.04      \\
\multicolumn{1}{l}{Joint-decoder}                          & - & - & \textbf{74.53}          & - & - & 62.30          & - &- & 63.10          & - & - & 74.27    &68.55      \\

\hline
\multicolumn{1}{l}{$S^3E^2$}                         & 69.08 & 64.55 & 66.74          & 59.43 & 46.23 & 52.01          & 61.06 & 56.44 & 58.66          & 71.08 & 63.13 & 66.87   &61.07       \\
\multicolumn{1}{l}{GTS-BERT}                        & 70.92 & 69.49 & 70.20          & 57.52 & 51.92 & 54.58          & 59.29 & 58.07 & 58.67          & 68.58 & 66.60 & 67.58   &62.76       \\
\multicolumn{1}{l}{EMC-GCN}                            & 71.85 & \textbf{72.12} & 71.98          & 61.46 & 55.56 & 58.32          & 59.89 & 61.05 & 60.38          & 65.08 & 71.66 & 68.18  &64.72        \\ 
\hline
\multicolumn{1}{l}{PT-GCN(Ours)}               &\textbf{76.62}  &71.96  &74.22 &\textbf{71.36}  &\textbf{56.08}  &\textbf{62.80} &\textbf{71.29}  &\textbf{63.26}  &\textbf{67.04}         &\textbf{74.03}  &\textbf{75.35} &\textbf{74.68} & \textbf{69.69}\\
\multicolumn{1}{l}{$\Delta$}                            & $\uparrow$\textbf{4.77} & - & -         & $\uparrow$\textbf{9.90} & $\uparrow$\textbf{0.52} & $\uparrow$\textbf{0.50}          & $\uparrow$\textbf{10.23} & $\uparrow$\textbf{2.21} & $\uparrow$\textbf{3.94}          & $\uparrow$\textbf{2.95} & $\uparrow$\textbf{3.69} & $\uparrow$\textbf{0.41} & $\uparrow$\textbf{1.14}          \\ 
\hline
\end{tabular}
}
\label{tab:mainresult-1}
\end{table*}

\begin{table*}[h]
\centering
\caption{Results on $\mathcal{D}_2$. $\dagger$ and $\ddagger$ denote that results are retrieved from \cite{xu-etal-2020-position} and \cite{Chen2022EnhancedMG}. }
\tabcolsep=2.0pt
\scalebox{1}{
\begin{tabular}{lllllllllllll|c}
\hline
\multicolumn{1}{c}{\multirow{2}{*}{\textbf{Model}}}
    & \multicolumn{3}{c}{14res}      & \multicolumn{3}{c}{14lap}      & \multicolumn{3}{c}{15res}           & \multicolumn{3}{c|}{16res}  & \multirow{2}{*}{Avg-F1}     \\ \cline{2-13} 
\multicolumn{1}{l}{}                                & P.    & R.        & F1             & P.    & R.    & F1             & P.    & R.    & F1             & P.    & R.    & F1    &{}         \\ \hline
\multicolumn{1}{l}{Peng-two-stage$^\dagger$}                  & 43.24 & 63.66 & 51.46          & 37.38 & 50.38 & 42.87          & 48.07 & 57.51 & 52.32          & 46.96 & 64.24 & 54.21    &50.22      \\ %

\multicolumn{1}{l}{BMRC$^\ddagger$}                            & 75.61 & 61.77 & 67.99          & \textbf{70.55} & 48.98 & 57.82          & 68.51 & 53.40 & 60.02          & 71.20 & 61.08 & 65.75   &62.90       \\%
\multicolumn{1}{l}{COM-MRC}                          & 75.46 & 68.91 & 72.01          & 62.35 & \textbf{58.16} & 60.17          & 68.35 & 61.24 & 64.53          & 71.55 & 71.59 & 71.57   &67.07       \\
\hline
\multicolumn{1}{l}{JET-BERT$^\dagger$}                        & 70.56 & 55.94 & 62.40          & 55.39 & 47.33 & 51.04          & 64.45 & 51.96 & 57.53          & 70.42 & 58.37 & 63.83     & 58.70    \\%
\multicolumn{1}{l}{BART-ABSA}                       & 65.52 & 64.99 & 65.25          & 61.41 & 56.19 & 58.69          & 59.14 & 59.38 & 59.26          & 66.60 & 68.68 & 67.62    &62.71      \\
\multicolumn{1}{l}{Span-ASTE}                  & 72.89 & 70.89 & 71.85          & 63.44 & 55.84 & 59.38          & 62.18 & 64.45 & 63.27 & 69.45 & 71.17 & 70.26    &66.19      \\ 
\multicolumn{1}{l}{RLI}                  &77.46 &71.97 &74.34    &63.32 &57.43 &60.96   &60.08 &\textbf{70.66} &65.41 &70.50 &\textbf{74.28} &72.34   &68.26       \\ 

\hline
\multicolumn{1}{l}{GTS-BERT$^\ddagger$}                        & 68.09 & 69.54 & 68.81          & 59.40 & 51.94 & 55.42          & 59.28 & 57.93 & 58.60          & 68.32 & 66.86 & 67.58      &62.60    \\%
\multicolumn{1}{l}{Dual-Encoder}                        & 67.95 & 71.23 & 69.55          & 62.12 & 56.38 & 59.11          & 58.55 & 60.00 & 59.27          & 70.65 & 70.23 & 70.44     &64.59     \\
\multicolumn{1}{l}{EMC-GCN}                          & 71.21 & 72.39 & 71.78          & 61.70 & 56.26 & 58.81          & 61.54 & 62.47 & 61.93          & 65.62 & 71.30 & 68.33    &65.21      \\
\multicolumn{1}{l}{BDTF}                          &75.53 &\textbf{73.24} &74.35     &68.94 &55.97 &61.74    &68.76 &63.71 &66.12    &71.44 &73.13 &72.27    &68.62      \\

\hline
\multicolumn{1}{l}{PT-GCN(Ours)}               &\textbf{77.85}  &72.13  & \textbf{74.88}    &67.69  &57.30  & \textbf{62.06} &\textbf{69.76}  &65.15  & \textbf{67.38}          &\textbf{74.39}  &71.21 & \textbf{72.76} &\textbf{69.27} \\ 
\multicolumn{1}{l}{$\Delta$}                            & $\uparrow$\textbf{0.39} & - & $\uparrow$\textbf{0.53}          & - & - & $\uparrow$\textbf{0.32}          & $\uparrow$\textbf{1.00} & - & $\uparrow$\textbf{1.26}          & $\uparrow$\textbf{2.84} & - & $\uparrow$\textbf{0.49}  &$\uparrow$\textbf{0.65}        \\ 
\hline
\end{tabular}
}

\label{tab:mainresult-2}
\end{table*}

\subsection{Baselines.}

\textbf{1) Pipeline} methods decompose the ASTE task: Peng-two-stage \cite{DBLP:conf/aaai/PengXBHLS20}, Peng+IOG  \cite{wu-etal-2020-grid} and IMN+IOG \cite{wu-etal-2020-grid};
or treat it as a multi-step machine reading comprehension task: BMRC \cite{DBLP:conf/aaai/ChenWLW21}, Dual-MRC \cite{DBLP:conf/aaai/MaoSYC21} and COM-MRC \cite{zhai-etal-2022-com}. 
\textbf{2) End-to-End} methods address the task with an encoder-docoder architecture: JET-BERT \cite{xu-etal-2020-position}, Joint-Decoder \cite{zhao-etal-2022-learning-cooperative}, BART-ABSA \cite{yan-etal-2021-unified}, Span-ASTE \cite{xu-etal-2021-learning} and RLI\cite{yu-etal-2023-making}.
\textbf{3) Table-filling} methods encoder the word relations into a two-dimensional table:
$S^3E^2$ \cite{chen-etal-2021-semantic}, GTS-BERT \cite{wu-etal-2020-grid}, EMC-GCN \cite{Chen2022EnhancedMG}, Dual-Encoder \cite{jing-etal-2021-seeking} and BDTF \cite{zhang-etal-2022-boundary}.


\subsection{Implementation.}
We use Bert-base-uncased\footnote{https://huggingface.co/bert-base-uncased} \cite{DBLP:conf/nips/VaswaniSPUJGKP17} as our encoder.
We train the model for 20 epochs with a batch size of 4, and the learning rate is set to $3\times 10^{-5}$.
In each epoch, we evaluate the training model on the development set and save the best one.
We set $d=768$ and $l=2$.
Without special instructions, the default settings for $k$ and $\alpha$ are 0.3 and 0.5, respectively.
All results (F1-score) are the average across five runs with different random seeds.
We perform all experiments on an NVIDIA GeForce RTX 3090.

\subsection{Main Results.}
The main results are shown in Table \ref{tab:mainresult-1} and Table \ref{tab:mainresult-2}, where the best results are highlighted in bold. 
$\alpha$ is set to 0.4 on $\mathcal{D}_1$-14lap and 0.6 on $\mathcal{D}_1$-16res.
It can be observed that our PT-GCN outperforms all the baselines.
Table \ref{tab:mainresult-1} shows the result on $\mathcal{D}_1$. 
Compared with the best baseline Join-decoder, the improvements in F1-score of our PT-GCN are 0.50\%, 3.94\%, and 0.41\% on 14lap, 15res, and 16res, respectively. But drop by 0.31\% on 14res.
The average improvement is 1.14\%.
Table \ref{tab:mainresult-2} shows the result on $\mathcal{D}_2$. 
Compared with the best baseline BDTF, the improvements in F1-score of our PT-GCN are 0.53\%, 0.32\%, 1.26\%, and 0.49\% on four datasets, respectively. The average improvement is 0.65\%.
The BDTF focuses on learning relation interactions but ignores the sentiment features, while our model can automatically learn the relation interactions of different sentiment polarities at the instance level.
The results demonstrate that our model achieves state-of-the-art performance.

Compared with other table-filling models, our model outperforms the best baseline EMC-GCN by an average of 4.06\% in F1-score.
Although EMC-GCN introduced additional syntactic information (such as part-of-speech and syntactic dependency) through a multi-channel GCN module, our model performs better without external knowledge.
The reason is that rather than using external linguistic knowledge, prompt learning can utilize the rich knowledge contained in the language model.
Compared with GTS-BERT, which simply uses the BERT representation for table filling, the improvement of our model is 6.67\% in F1-score.

\begin{table}[t]\small
\centering
\caption{Analysis of wrong predictions on $\mathcal{D}_2$.}
\tabcolsep=1.3pt
\scalebox{1}{
\begin{tabular}{lllllllll}
\hline
\multicolumn{1}{c}{\multirow{2}{*}{Model}} & \multicolumn{2}{c}{14res} & \multicolumn{2}{c}{14lap} & \multicolumn{2}{c}{15res} & \multicolumn{2}{c}{16res} \\ \cline{2-9} 
\multicolumn{1}{c}{}                       & $\emph{E.}$         & $\emph{S.}$        & $\emph{E.}$         & $\emph{S.}$        & $\emph{E.}$         & $\emph{S.}$        & $\emph{E.}$        & $\emph{S.}$        \\ \hline
GTS                                      & $10.16$        & $8.37$       & $11.75$        & $13.15$       & $14.37$        & $8.54$       & $9.21$       & $8.66$       \\
BDTF                             & $6.77$        & $5.52$       &$6.03$        & $12.06$       & $10.50$        & $7.22$       & $5.34$       & $7.71$       \\%
Ours      & $6.46$        & $5.11$       & $6.27$        & $11.51$       & $10.44$        & $7.04$       & $4.93$       & $7.20$       \\
$\Delta$      & $\downarrow0.31$        & $\downarrow0.41$       & $-$         & $\downarrow1.09$       & $\downarrow0.06$        & $\downarrow0.18$       & $\downarrow0.41$       & $\downarrow0.51$       \\ 
\hline
\end{tabular}
}
\label{tab:result2}
\end{table}

\begin{table*}[th]
\caption{Ablation study on $\mathcal{D}_2$. The evaluation metrics are F1-score and single sentence inference time cost (ms).}
\centering
\tabcolsep=3.0pt
\begin{tabular}{lccccccccc|cccc}
\hline
\multirow{2}{*}{Models} & \multirow{2}{*}{\#Params} & \multicolumn{2}{c}{14res} & \multicolumn{2}{c}{14lap} & \multicolumn{2}{c}{15res} & \multicolumn{2}{c|}{16res} & \multicolumn{2}{c}{Avg.} & \multicolumn{2}{c}{Avg.$\Delta$} \\ \cline{3-14} 
                        &                           & F1          & ms.       & F1          & ms.       & F1          & ms.       & F1           & ms.       & F1          & ms.      & F1            & ms.(\%)            \\ \hline
PT-GCN                  & 237.6M                    & \textbf{74.88}       & 36.23       & \textbf{62.06}       & 37.79       & \textbf{67.38}       & 38.50       & \textbf{72.76}        & 38.76       & \textbf{69.27}       & 37.82      & -            &   $\uparrow87.88\%$       \\
\quad No-senti                & 237.6M                    & 72.11       &   36.49          & 61.63       &   37.44          & 64.96       &   38.23          & 72.00        &   38.82          &  67.68           &  37.75      &  $\downarrow1.59$             &  $\uparrow87.53\%$       \\
\quad Single                  & 152.2M                          & 71.85       &    32.11         & 60.35       & 26.28            & 64.40        &   26.08      & 71.41        &   34.84            &67.00             &   29.83       &  $\downarrow2.27$             &  $\uparrow48.17\%$                \\
\quad None                    & 110.8M                    & 68.49       & \textbf{17.79}       & 55.23       & \textbf{20.31}       & 61.32       & \textbf{22.02}       & 66.42        & \textbf{20.41}       & 62.87       & \textbf{20.13}      &   $\downarrow6.40$            & -                \\
\hline
\end{tabular}
\label{tab:Ablation}
\end{table*}

To reveal the improvements in our model, we conduct an experiment to analyze the wrong predictions.
As shown in Table~\ref{tab:result2}, where \emph{E.} denotes entity errors and \emph{S.} denotes sentiment errors.
Entity error means a triplet with a correct sentiment term and wrong aspect terms or opinion terms.
By contrast, sentiment errors are caused by the wrong sentiment, while the other two entities are correct.
Compared with the best baseline BDTF on $\mathcal{D}_2$, our model's entity error rate has decreased by an average of 0.14\%, and the sentiment error rate has decreased by an average of 0.56\%.
It can be observed that our model shows a more significant decrease in sentiment error rate.
This proves that our model has a stronger learning ability for sentiment relations.

\subsection{Ablation Study.}
To further reveal the effectiveness of the tri-channel GCN module, we conduct an ablation study on $\mathcal{D}_2$.
The results are shown in F1-score in Table~\ref{tab:Ablation}. 
In the \textbf{No-Senti} setting, we remove the sentiment polarity description (the words ``positive'', ``negative'', and ``neutral'') from the prompt template in Figure \ref{fig_main}.
The results show that the performance will drop by an average of 1.59\%.
This is because the triple channel GCN benefits from the ability to learn from multiple sentiment perspectives.

In the \textbf{Single} setting, instead of using a template with six sentiment mask slots, we use a template with only two mask slots, like ``\emph{aspect [MASK], opinion [MASK].}''.
The triple channel graph is modified to only a single graph without sentiment perception.
The model performance decreases by 2.27\%.
Compared to the No-Senti setting, the Single setting only slightly decreased more by 0.31\%.
This reveals that it is the sentiment indication in the template, not the number of slots, that plays an important role in the model's performance.

In the \textbf{None} setting, we remove the Section \ref{tabletograph} Table-to-Graph Transformation step and use the table $C$ directly as inputs to the table-decoding module.
The results show that the performance will drop by an average of 6.40\%.
This reveals that the Table-to-Graph step is beneficial for learning word relations.

\subsection{Computing Efficiency.}
As shown in Table~\ref{tab:Ablation}, we evaluate our model's computing efficiency by the number of parameters and single sentence inference time cost.
In the \textbf{None} setting, the model retains the backbone and has the lowest inference time cost, but the performance of the model decreases significantly.
Comparing \textbf{Singe} with \textbf{None}, the model performance has improved and the time cost has slightly increased.
In the \textbf{No-senti} setting, compared to the full parameter, there is no change in parameter quantity and time cost, as they only have different inputs.
%
Our full model has an average inference time of 37.82 ms, which is close to that of  Single, while being less than double that of None. Thus balance between performance and time cost is achieved.

\subsection{Analysis of Coefficient.}
\begin{figure}[t]
\begin{minipage}[b]{1\linewidth}
  \centering
  \centerline{\includegraphics[width=0.99\textwidth]{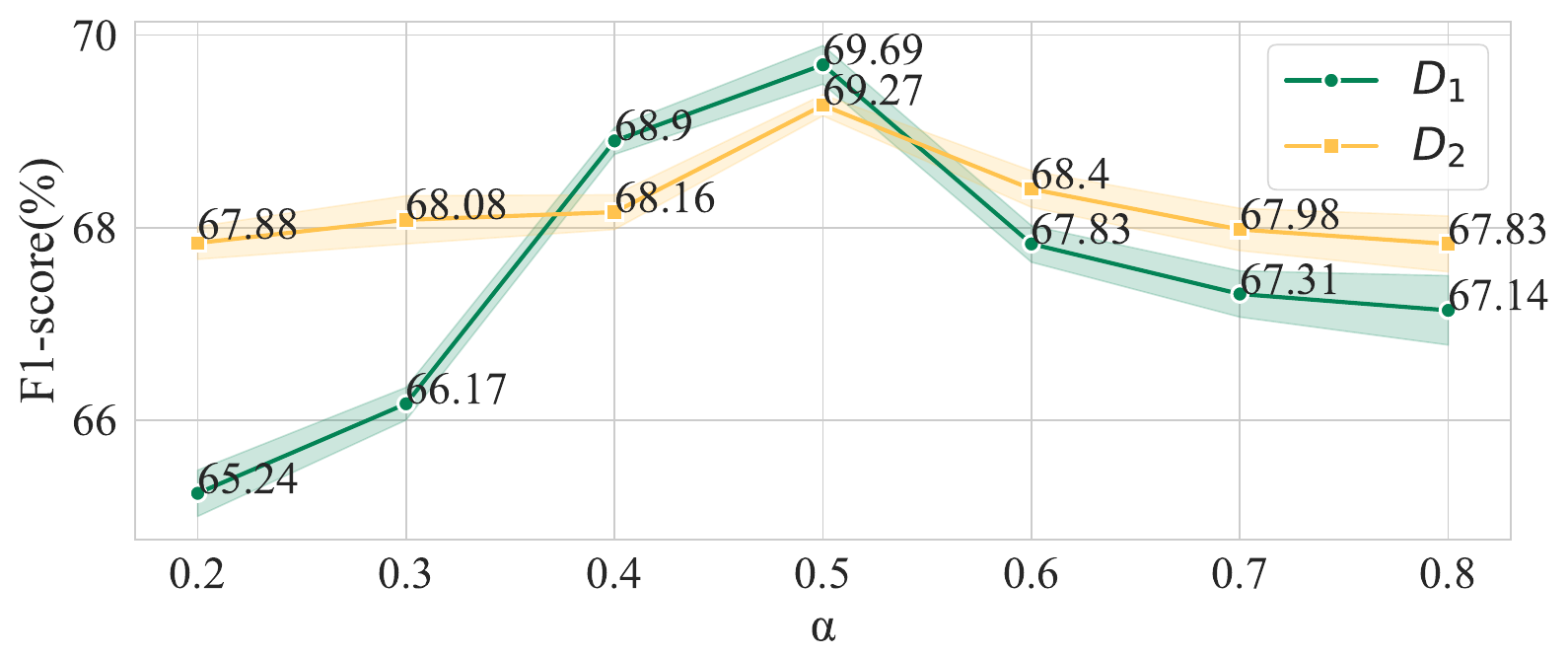}}
  \vspace{-0.2cm}
  \centerline{(a) Effect of different $\alpha$ while set $k=0.3$.}\label{hxa}\medskip
\end{minipage}
\begin{minipage}[b]{1\linewidth}
  \centering
  \centerline{\includegraphics[width=0.98\textwidth]{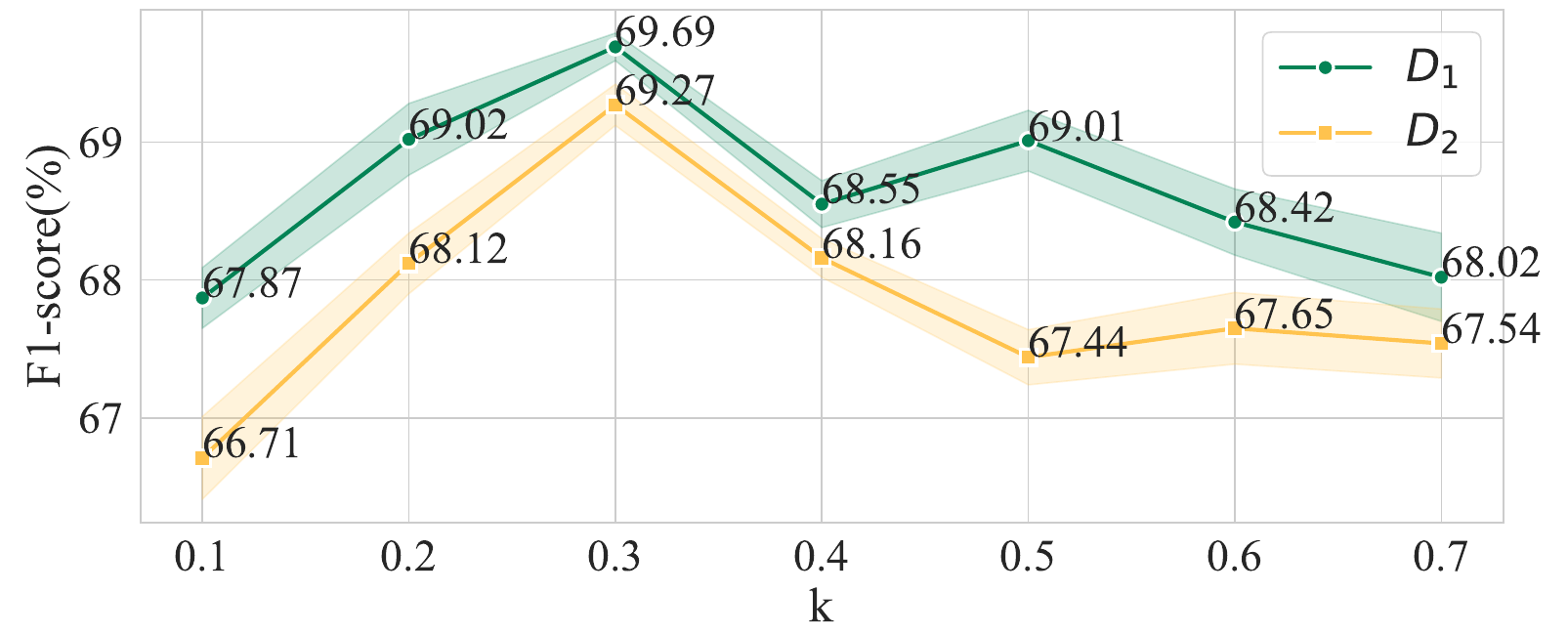}}
  \vspace{-0.2cm}
  \centerline{(b) Effect of different $k$ while set $\alpha=0.5$.}\label{hxb}\medskip
\end{minipage}
\vspace{-0.75cm}
\caption{Effects of $\alpha$ and $k$.}
\label{fig3}
\end{figure}

To further investigate the effect of $\alpha$ and $k$, we conduct two experiments on $\mathcal{D}_1$ and $\mathcal{D}_2$.
We use the average F1-score on the four subdatasets as metrics.
We first set $k=0.3$ and study the coefficient $\alpha$ from value 0.2 to 0.8.
As shown in Figure~\ref{fig3}(a), the best value of $\alpha$ is 0.5 for both $\mathcal{D}_1$ and $\mathcal{D}_2$.
We also study the span pruning threshold $k$ from value 0.1 to 0.7 while set $\alpha=0.5$.
The value of $k$ controls the number of candidate entities in the \ref{tabledecoding} Table Decoding step, and the larger the value, the greater the number of entities to classify.
An excessive $k$ value can cause noise, while smaller values may cause candidates missing.
As shown in Figure~\ref{fig3}(b), the best value of $k$ is 0.3.

\begin{table*}[t]
\centering
\caption{Comparison results for AESC and ASOE tasks. $\dagger$ denotes that results are retrieved from \cite{DBLP:conf/aaai/MaoSYC21}.}
\scalebox{1}{
\begin{tabular}{lllllllll|ll}
\hline
\multicolumn{1}{c}{\multirow{2}{*}{Model}} 
    & \multicolumn{2}{c}{14res} & \multicolumn{2}{c}{14lap} & \multicolumn{2}{c}{15res} & \multicolumn{2}{c}{16res} & \multicolumn{2}{c}{Avg-F1} \\ \cline{2-11} 
    & AESC         & ASOE        & AESC         & ASOE        & AESC         & ASOE        & AESC        & ASOE   & AESC        & ASOE        \\ \hline
\multicolumn{1}{c}{CMLA+$^\dagger$}                                      & 70.62        & 48.95       & 56.90        & 44.10       & 53.60        & 44.60       & 61.20       & 50.00  & 60.58       & 46.91      \\
\multicolumn{1}{c}{Peng-two-stage$^\dagger$}                             & 74.19        & 56.10       & 62.34        & 53.85       & 65.79        & 56.23       & 71.73       & 60.04  & 68.51       & 56.56      \\%
\multicolumn{1}{c}{Dual-MRC$^\dagger$}                                   & 76.57        & 74.93       & 64.59        & 63.37       & 65.14        & 64.97       & 70.84       & 75.71   & 69.29       & 69.75    \\%

\multicolumn{1}{c}{BART-ABSA}                                  & 78.47        & 77.68       & 68.17        & 66.11       & 69.95        & 67.98       & 75.69       & 77.38   & 73.07       & 72.29     \\ 
\hline
\multicolumn{1}{c}{PT-GCN}      & \textbf{81.43}        & \textbf{78.90}    & \textbf{70.33}     & \textbf{69.95}     & \textbf{71.97}     & \textbf{69.31}      & \textbf{81.26}     & \textbf{79.99}   & \textbf{76.25}     & \textbf{74.54}  \\
\hline
\end{tabular}
}
\label{tab:subtask}
\end{table*}

\subsection{Performance on Subtasks.}
There are several subtasks of the ASTE task.
In this paper, we evaluate two typical of them: Aspect Term Extraction and Sentiment Classification (AESC) and Aspect-Opinion Pair Extraction (AOPE).
AESC aims at extracting all the aspect terms as well as the corresponding sentiment polarities simultaneously.
ASOE aims at extracting all the correct pairs of aspect-opinion terms from a sentence.
In the AESC setting, we directly extract the pair (aspect, sentiment) from the triplets.
In the ASOE setting, we remove the sentiment classification step.
Following previous studies, we evaluate our model on $\mathcal{D}_1$ in F1-score.
The comparison results are shown in Table~\ref{tab:subtask}, where the best results are highlighted in bold.
Our model outperforms the best pipeline model BART-ABSA by an average of 2.18\% and 1.25\% on the AESC and ASOE tasks, respectively.
The overall improvement indicates that PT-GCN is effective not only for the ASTE task but also for AESC and ASOE tasks.

\subsection{Prompt Score Visualization.}
\label{visualization}
\begin{figure}[t]
\begin{minipage}[b]{.48\linewidth}
  \centering
  \centerline{\includegraphics[width=4.5cm]{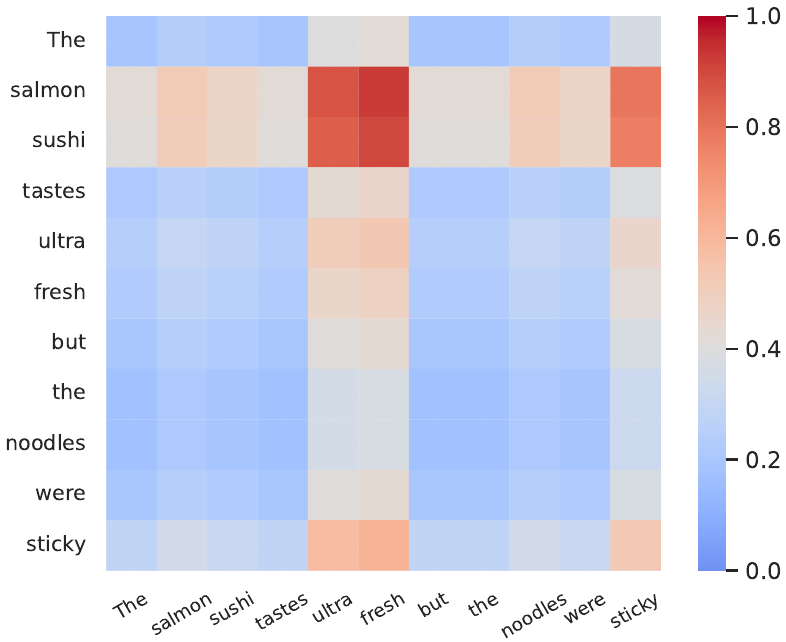}}
  \centerline{(a) Positive}\label{hba}\medskip
\end{minipage}
\begin{minipage}[b]{.48\linewidth}
  \centering
  \centerline{\includegraphics[width=4.5cm]{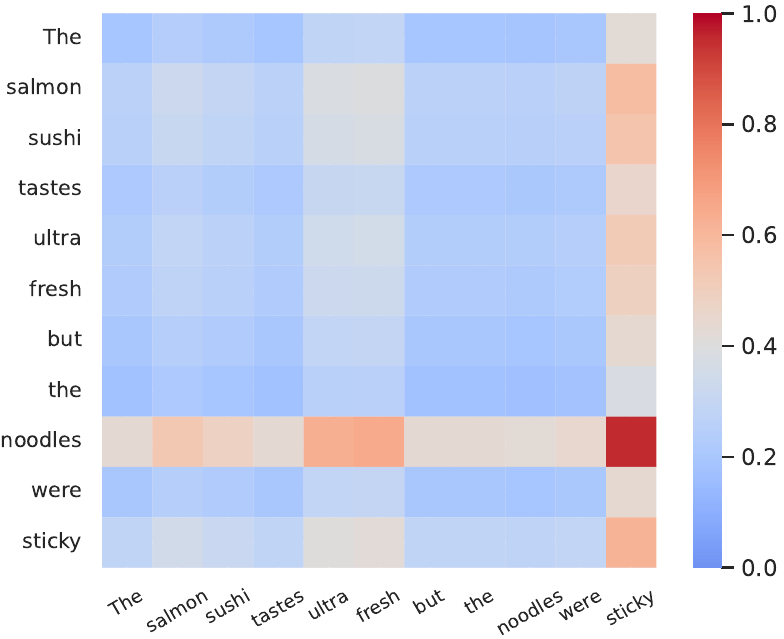}}
  \centerline{(b) Negative}\medskip
\end{minipage}
\begin{minipage}[b]{.48\linewidth}
  \centering
  \centerline{\includegraphics[width=4.5cm]{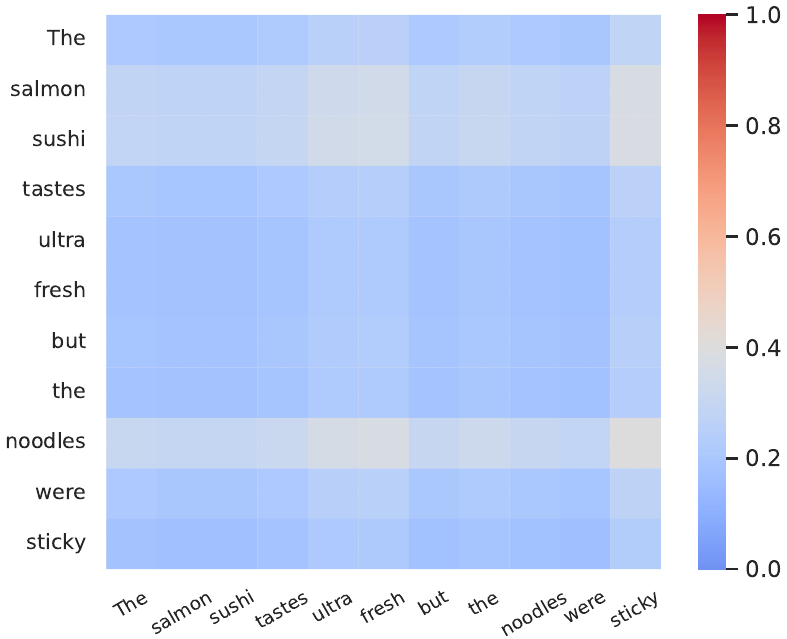}}
  \centerline{(c) Neutral}\medskip
\end{minipage}
\hfill
\begin{minipage}[b]{0.48\linewidth}
  \centering
  \centerline{\includegraphics[width=4.0cm]{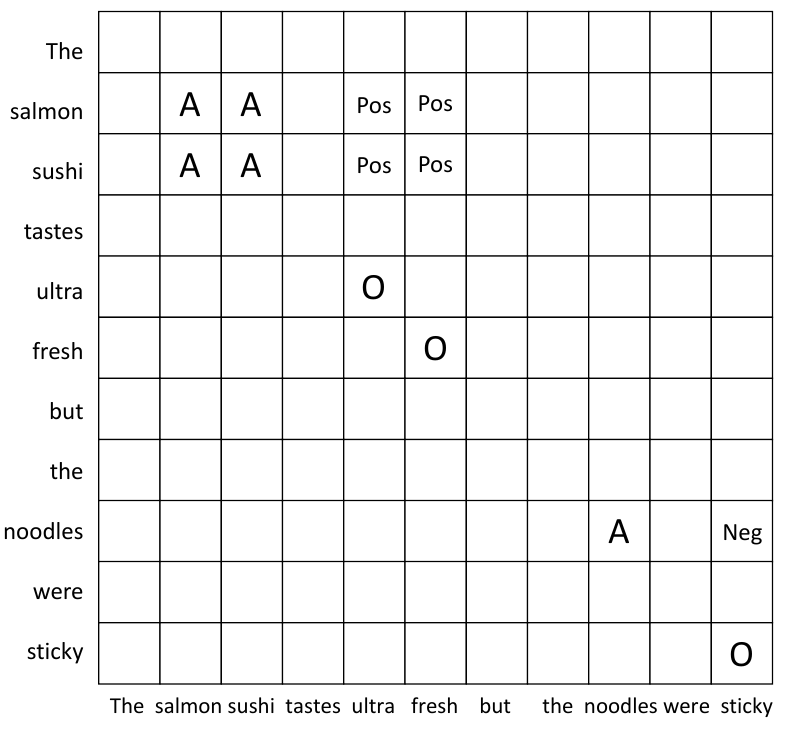}}
  \centerline{(d) Ground truth}\medskip
\end{minipage}
\caption{Heatmaps for different sentiment polarities and the corresponding ground truth.}

\label{fig:5}
\end{figure}

We use the case in Figure \ref{fig:1} as an example input.
To further investigate the role of prompt attention in the framework, we visualize the similarity heat maps of the three polarities, which are shown in Figure \ref{fig:5}.
It is worth noting that what we show here is the product of probabilities, not the visualization of the final result.
For each map, the vertical axis represents aspect probability $P^A$ and the horizontal axis represents opinion probability $P^O$, which are calculated in Eq. \ref{3.2}.
For each node $n_{ij}$ in the map, it is calculated by: $n_{ij} = \sqrt{p^a_i(p^o_j)^T} $.
We believe that $n_{ij}$ can somewhat indicate the aspect-opinion pairs.
For example, in the heat map (a), it can be easily observed that the darkest grids correspond to ``salmon sushi'' on the column and the ``ultra fresh'' on the horizontal.
By comparing (a) with ground truth (d), we find that $g_{ij}$ can indeed indicate the positions of golden pairs.
This conclusion can also be drawn in (b) and (c).
This proves that prompt attention plays a positive role in the entire model's performance.

\vspace{-0.3cm}

\section{Conclusion}
In this paper, we propose a novel PT-GCN approach to address the ASTE task.
Previous table-filling models simply clarify word-level relations in each individual cell but ignore the deep interaction between neighbor cells.
To accurately learn word relations, we propose a table-to-graph transformation method that does not need to rely on external knowledge.
Specifically, we propose a prompt attention mechanism to highlight the term words.
By treating the prompt attention score as edge weight and table cells as nodes, we transform the table into a grid-like target-aware graph.
Furthermore, we propose a parallel GCN module to fully utilize the information from different sentiment perspectives.
Extensive experiments show that our approach outperforms all baseline methods not only on the ASTE task but also on two subtasks.
The advantage of our model comes from a more potent ability to mine sentiment knowledge.
Detailed analysis demonstrates the effectiveness of each module within the overall framework.
In future work, we will delve into the different roles of the three channels in the parallel convolution module.

\section*{Acknowledgments}
This work is supported by Pilot Projects of Chinese Academy of Sciences (No.XDC02030400 and No.E3C0011), NSFC through grants 62322202, U21B2027, 62162037, U23A20388 and 62266028, Beijing Natural Science Foundation through grant 4222030, Yunnan Provincial Major Science and Technology Special Plan Projects through grants 202302AD080003, 202202AD080003 and 202303AP140008, General Projects of  Basic Research in Yunnan Province through grant 202301AS070047, 202301AT070471, and the Fundamental Research Funds for the Central Universities. 
Philip S. Yu was supported in part by NSF under grant III-2106758.

\bibliographystyle{unsrt}
\bibliography{ref.bib}

\begin{thebibliography}{10}

\bibitem{2014Sentiment}
Alvaro Ortigosa, Jos{\'e}~M. Mart{\'i}n, and Rosa~M. Carro.
\newblock Sentiment analysis in facebook and its application to e-learning.
\newblock {\em Comput. Hum. Behav.}, 31:527--541, 2014.

\bibitem{pontiki-etal-2014-semeval}
Maria Pontiki, Dimitris Galanis, John Pavlopoulos, Harris Papageorgiou, Ion
  Androutsopoulos, and Suresh Manandhar.
\newblock {S}em{E}val-2014 task 4: Aspect based sentiment analysis.
\newblock In {\em {S}em{E}val 2014}, pages 27--35.

\bibitem{zhao2023rdgcn}
Xusheng Zhao, Hao Peng, Qiong Dai, Xu~Bai, Huailiang Peng, Yanbing Liu,
  Qinglang Guo, and Philip~S Yu.
\newblock Rdgcn: Reinforced dependency graph convolutional network for
  aspect-based sentiment analysis.
\newblock {\em ACM WSDM 2024}.

\bibitem{2020Target}
H.~Wan, Y.~Yang, J.~Du, Y.~Liu, and J.~Z. Pan.
\newblock Target-aspect-sentiment joint detection for aspect-based sentiment
  analysis.
\newblock {\em AAAI}, 34(5):9122--9129, 2020.

\bibitem{DBLP:conf/aaai/PengXBHLS20}
Haiyun Peng, Lu~Xu, Lidong Bing, Fei Huang, Wei Lu, and Luo Si.
\newblock Knowing what, how and why: {A} near complete solution for
  aspect-based sentiment analysis.
\newblock In {\em AAAI}, pages 8600--8607, 2020.

\bibitem{xu-etal-2021-learning}
Lu~Xu, Yew~Ken Chia, and Lidong Bing.
\newblock Learning span-level interactions for aspect sentiment triplet
  extraction.
\newblock In {\em ACL-IJCNLP}, pages 4755--4766, 2021.

\bibitem{DBLP:conf/aaai/MaoSYC21}
Yue Mao, Yi~Shen, Chao Yu, and Longjun Cai.
\newblock A joint training dual-mrc framework for aspect based sentiment
  analysis.
\newblock In {\em AAAI 2021}, pages 13543--13551.

\bibitem{xu-etal-2020-position}
Lu~Xu, Hao Li, Wei Lu, and Lidong Bing.
\newblock Position-aware tagging for aspect sentiment triplet extraction.
\newblock In {\em EMNLP}, pages 2339--2349, 2020.

\bibitem{yan-etal-2021-unified}
Hang Yan, Junqi Dai, Tuo Ji, Xipeng Qiu, and Zheng Zhang.
\newblock A unified generative framework for aspect-based sentiment analysis.
\newblock In {\em ACL-IJCNLP 2021}.

\bibitem{zhang-etal-2021-towards-generative}
Wenxuan Zhang, Xin Li, Yang Deng, Lidong Bing, and Wai Lam.
\newblock Towards generative aspect-based sentiment analysis.
\newblock In {\em ACL-IJCNLP}, pages 504--510, 2021.

\bibitem{2016Table}
P.~Gupta, H~Schütze, and B.~Andrassy.
\newblock Table filling multi-task recurrent neural network for joint entity
  and relation extraction.
\newblock In {\em ICCL}, 2016.

\bibitem{wu-etal-2020-grid}
Zhen Wu, Chengcan Ying, Fei Zhao, Zhifang Fan, Xinyu Dai, and Rui Xia.
\newblock Grid tagging scheme for aspect-oriented fine-grained opinion
  extraction.
\newblock In {\em Findings of EMNLP}, pages 2576--2585, 2020.

\bibitem{Chen2022EnhancedMG}
Hao Chen, Zepeng Zhai, Fangxiang Feng, Ruifan Li, and Xiaojie Wang.
\newblock Enhanced multi-channel graph convolutional network for aspect
  sentiment triplet extraction.
\newblock In {\em ACL}, pages 2974--2985, 2022.

\bibitem{zhang-etal-2022-boundary}
Yice Zhang, Yifan Yang, Yihui Li, Bin Liang, Shiwei Chen, Yixue Dang, Min Yang,
  and Ruifeng Xu.
\newblock Boundary-driven table-filling for aspect sentiment triplet
  extraction.
\newblock In {\em EMNLP}, pages 6485--6498, 2022.

\bibitem{DBLP:conf/nips/VaswaniSPUJGKP17}
Ashish Vaswani, Noam Shazeer, Niki Parmar, Jakob Uszkoreit, and et~al.
\newblock Attention is all you need.
\newblock In {\em NIPS}, pages 5998--6008, 2017.

\bibitem{jing-etal-2021-seeking}
Hongjiang Jing, Zuchao Li, Hai Zhao, and Shu Jiang.
\newblock Seeking common but distinguishing difference, a joint aspect-based
  sentiment analysis model.
\newblock In {\em EMNLP 2021}.

\bibitem{chen-qian-2019-transfer}
Zhuang Chen and Tieyun Qian.
\newblock Transfer capsule network for aspect level sentiment classification.
\newblock In {\em ACL}, pages 547--556, 2019.

\bibitem{DBLP:conf/aaai/ChenWLW21}
Shaowei Chen, Yu~Wang, Jie Liu, and Yuelin Wang.
\newblock Bidirectional machine reading comprehension for aspect sentiment
  triplet extraction.
\newblock {\em AAAI 2021}.

\bibitem{Brown2020LanguageMA}
Tom~B. Brown, Benjamin Mann, Nick Ryder, and et~al.
\newblock Language models are few-shot learners.
\newblock In {\em NeurIPS 2020}.

\bibitem{petroni-etal-2019-language}
Fabio Petroni, Tim Rockt{\"a}schel, Sebastian Riedel, Patrick Lewis, Anton
  Bakhtin, Yuxiang Wu, and Alexander Miller.
\newblock Language models as knowledge bases?
\newblock In {\em EMNLP-IJCNLP}, pages 2463--2473, 2019.

\bibitem{hu-etal-2022-knowledgeable}
Shengding Hu, Ning Ding, Huadong Wang, and et~al.
\newblock Knowledgeable prompt-tuning: Incorporating knowledge into prompt
  verbalizer for text classification.
\newblock In {\em ACL}, pages 2225--2240, 2022.

\bibitem{qi-etal-2022-enhancing}
Kunxun Qi, Hai Wan, Jianfeng Du, and Haolan Chen.
\newblock Enhancing cross-lingual natural language inference by prompt-learning
  from cross-lingual templates.
\newblock In {\em ACL}, pages 1910--1923, 2022.

\bibitem{wu-shi-2022-adversarial}
Hui Wu and Xiaodong Shi.
\newblock Adversarial soft prompt tuning for cross-domain sentiment analysis.
\newblock In {\em ACL}, pages 2438--2447, 2022.

\bibitem{li2021sentiprompt}
Chengxi Li, Feiyu Gao, Jiajun Bu, and et~al.
\newblock Sentiprompt: Sentiment knowledge enhanced prompt-tuning for
  aspect-based sentiment analysis.
\newblock {\em CoRR}, abs/2109.08306, 2021.

\bibitem{socher-etal-2013-recursive}
Richard Socher, Alex Perelygin, Jean Wu, Jason Chuang, Christopher~D. Manning,
  Andrew Ng, and Christopher Potts.
\newblock Recursive deep models for semantic compositionality over a sentiment
  treebank.
\newblock In {\em EMNLP}, pages 1631--1642, 2013.

\bibitem{kipf2017semisupervised}
Thomas~N. Kipf and Max Welling.
\newblock Semi-supervised classification with graph convolutional networks.
\newblock In {\em ICLR}, 2017.

\bibitem{zhai-etal-2022-com}
Zepeng Zhai, Hao Chen, Fangxiang Feng, Ruifan Li, and Xiaojie Wang.
\newblock {COM}-{MRC}: A {CO}ntext-masked machine reading comprehension
  framework for aspect sentiment triplet extraction.
\newblock In {\em EMNLP 2022}.

\bibitem{zhao-etal-2022-learning-cooperative}
Shiman Zhao, Wei Chen, and Tengjiao Wang.
\newblock Learning cooperative interactions for multi-overlap aspect sentiment
  triplet extraction.
\newblock In {\em EMNLP 2022}.

\bibitem{yu-etal-2023-making}
Guoxin Yu, Lemao Liu, Haiyun Jiang, Shuming Shi, and Xiang Ao.
\newblock Making better use of training corpus: Retrieval-based aspect
  sentiment triplet extraction via label interpolation.
\newblock In {\em ACL}, pages 4914--4927, 2023.

\bibitem{chen-etal-2021-semantic}
Zhexue Chen, Hong Huang, Bang Liu, Xuanhua Shi, and Hai Jin.
\newblock Semantic and syntactic enhanced aspect sentiment triplet extraction.
\newblock In {\em Findings of ACL-IJCNLP}, pages 1474--1483, 2021.

\end{thebibliography}

\end{document}